\documentclass{article}
\usepackage{spconf,amsmath,graphicx}
\usepackage{xcolor}
\usepackage{epsfig}
\usepackage{algorithm}
\usepackage[noend]{algpseudocode}
\usepackage{mathtools}

\title{Learning to identify image manipulations in scientific publications }
\name{Ghazal Mazaheri,~ Kevin Urrutia Avila,~Amit K. Roy-Chowdhury}
\address{University of California, Riverside, CA 92521}

\begin{document}
%
\maketitle
\begin{abstract}

Adherence to scientific community standards ensures objectivity, clarity, reproducibility, and helps prevent bias, fabrication, falsification, and plagiarism. To help scientific integrity officers and journal/publisher reviewers monitor if researchers stick with these standards, it is important to have a solid procedure to detect duplication as one of the most frequent types of manipulation in scientific papers. Images in scientific papers are used to support the experimental description and the discussion of the findings. Therefore, in this work we focus on detecting the duplications in images as one of the most important parts of a scientific paper. We propose a framework that combines image processing and deep learning methods to classify images in the articles as duplicated or unduplicated ones. We show that our method leads to a 90\% accuracy rate of detecting duplicated images, a $\sim {13\%}$ improvement in detection accuracy in comparison to other manipulation detection methods. We also show how effective the pre-processing steps are by comparing our method to other state-of-art manipulation detectors which lack these steps. 

\end{abstract}
\section{Introduction}
\label{sec:intro}

Publication record is one the most important work quality measurements for research scientists. 
Results in publications can lead to new drugs, products, treatment options, etc. and can have a huge impact on society. Scientists build on the results of previous publications. Thus, it is very important to protect research publications from plagiarism or duplication. In this paper, we focus on the problem of identifying duplication in images in research publications.

Figures and plots are key sections within any research paper as they convey important information regarding approach or results. When we talk about misconduct in research writing or publishing, we often think of plagiarism or duplicate publications. Another common problem under the umbrella of misconduct is that of image manipulation or duplication. Inappropriate image duplications in scientific papers occur when duplicate figure panels or parts of panels are present multiple times within the same paper, while representing different experiments. These can be the result of honest error, or done deliberately.


Based on the research by Bik \cite{bik}, a total of 20,621 research papers containing the search term “Western blot” from 40 different journals and 14 publishers were examined for inappropriate duplications of images. Of these, 8,138 (39.8\%) were published by a single journal (PLoS One) in 2013 and 2014; the other 12,483 (60.5\%) papers were published in 39 journals spanning the years 1995 to 2014 . Overall, 782 (3.8\%) of these papers were found to include at least one figure containing inappropriate duplications.

\begin{figure}[ht]
\includegraphics[width=\linewidth]{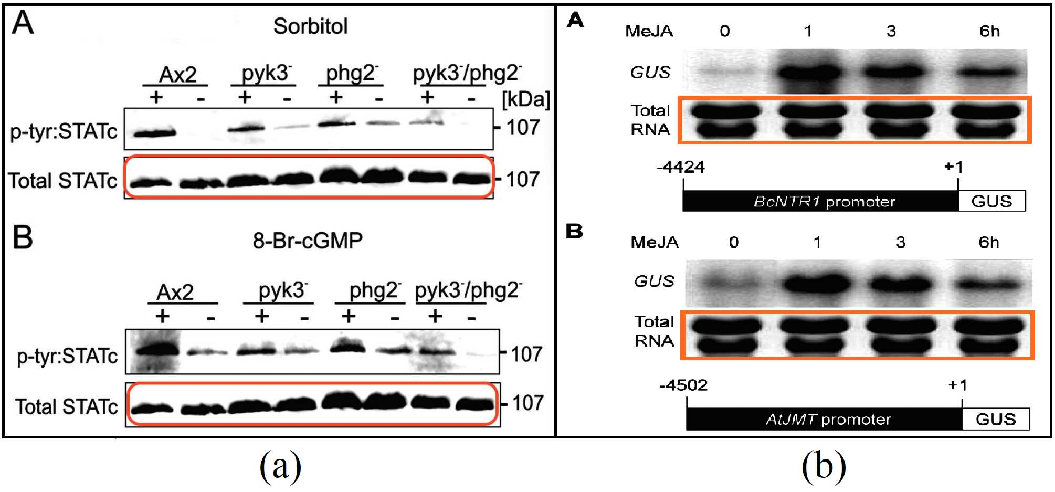}
\caption{(a) and (b) are two figures extracted from the articles \cite{Vu,Seo} containing image duplications. Red bounding boxes in these figures show the copy-paste regions.} \label{fig1}
\end{figure}

Manipulation in biomedical publications can affect the medical area negatively and has irreparable effects. Therefore, it is necessary to have a solid framework to detect manipulation within this specific research area. In this work, we focus on the problem of image duplication detection in the biomedical papers using the dataset provided by Bik \cite{bik}. Fig. \ref{fig1} demonstrates an example of image duplication in a biomedical research paper. As we can see, the same figure is used to represent the results of two different experiments.

\begin{figure*}[t]
\includegraphics[width=\linewidth]{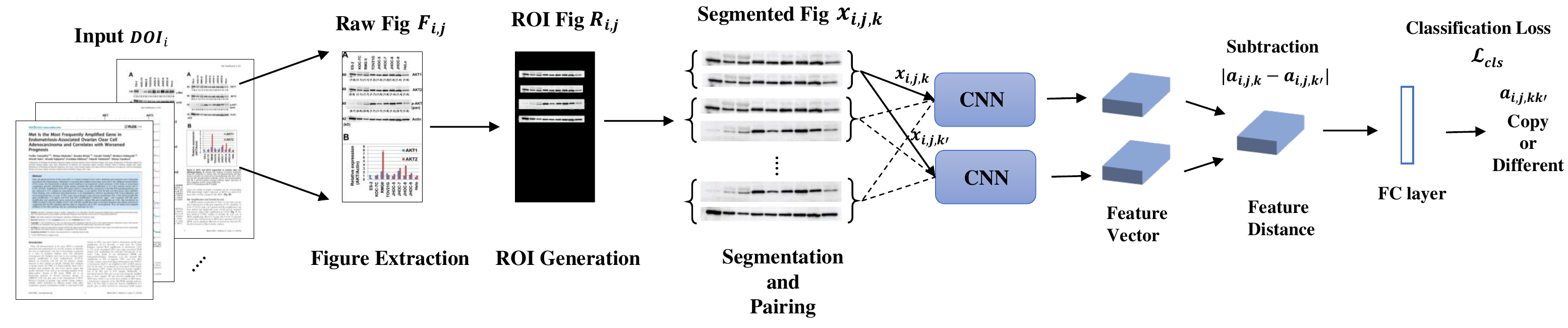}
\caption{This figure represents our proposed approach for image duplication detection in scientific articles. The proposed method includes pre-processing phase (figure extraction and ROI generation steps) and detection step.} \label{fig2}
\end{figure*}

\textbf{Framework Overview.} A  pictorial  flow  of our image duplication detection framework  is  presented  in  Fig. \ref{fig2}. The proposed method utilizes three main steps.\\ 
1. Extracting the figures out of papers.\\ 
2. Region of Interest (ROI) generation.\\ 
3. Duplication detection using Siamese network.\\
In the first step, we use Digital Object Identifier (DOI) of the clean and dublicated papers along with figure numbers from provided \cite{bik} to extract the images and prepare the dataset. In the next step, we extract the regions of interest excluding texts, graphs and charts. These two steps are necessary to prepare appropriate dataset for detection.  After two steps of data pre-processing, we segment out the regions of interest, pair the images and feed the image pairs to Siamese network for duplication detection.

\textbf{Main contributions.}
We propose a new approach to detect image duplication in scientific articles building upon the above three steps of image extraction, region of interest generation and copy-paste detection using Siamese network. Our method leads to $\sim {13\%}$  of improvement in detection accuracy in comparison to other manipulation detection methods. We also show how effective the pre-processing steps are by comparing our method to other state-of-art manipulation detectors where the raw figures extracted out of papers are fed to them.

\section{Related Work}
\label{sec:format}
Research paper authors disrespect scientific integrity rules when they manipulate the real information with the purpose of deceiving the reader, concealing data features, or fabricating whole or parts of an image. There have been several proposals to understand the extent of the misconduct problem in science. Fanelli \cite{Faneli} collected data from surveys and meta-analysis to answer how many scientists falsify or fabricate their scientific results. To understand the correlation between journal impact factor and data duplication, Oksvold \cite{ok} selected 120 random articles from three different journals related to cancer, with different impact factors. With his analysis, he found that 25\% of journal articles of impact factor less than five or greater than twenty contain data duplication. 

Knowledge of computer science can help detection of research misconduct. Authors in \cite{Acuna} have built a pipeline to analyze potential inappropriate reuse of figures in the biological sciences literature. Their method is a combination of image patche classification into biological or non-biological ones and image copy-move detection. 
Bucci \cite{bucci} introduce a software pipeline to detect some of the most diffuse misbehaviours. They discovered 6\% of published papers in their dataset contain manipulated images.

Inspired by the success of deep neural networks in different visual recognition tasks in computer vision, deep learning-based approaches have been popular choices for image forgery detection. Therefore, we propose a method using deep neural networks on top of image pre-proccesing methods to detect image duplication.

\section{Methodology}
\label{sec:pagestyle}

In this section, we present our framework for image duplication detection.  We start with a formal description of the problem statement followed by the three main steps of our approach including figure extraction, region of interest generation and duplication detection.
\vspace{-3mm}
\subsection{Problem Statement}
Consider we have a dataset of tuples from the image pre-processing step:
$\mathcal{X}=\{(x_{i,j,k},x_{i,j,k'},y_{i,j,kk'})\}$, where $x_{i,j,k}$ and $x_{i,j,k'}$ are $k_{th}$ and $k'_{th}$ segmented figures from $j_{th}$ ROI figure $R_{i,j}$.  ROI figure $R_{i,j}$ is output of ROI generation step from $j_{th}$ raw image $F_{i,j}$ which is extracted out of $i_{th}$ scientific article $DOI_i$. $y_{i,j,kk'} \in \{0,1\}$ is an indicator whether $x_{i,j,k}$ and $x_{i,j,k'}$ are copied or not. Given such a dataset, our main goal is to learn a model that would be able to classify a pair of test images to be either copied or not.

\vspace{-3mm}
\subsection{Figure Extraction}
Research publications are presented in different formats, and the images are also formatted in different ways within the text, thus making it essential for available data to be organized for analysis. This process typically includes converting data from raw form into another format to allow for more convenient consumption and organization. In our work, this data wrangling step is part of the pre-processing strategy to prepare the appropriate dataset for detection of duplication or otherwise. In this step, we use DOI of clean and duplicated papers along with figure numbers from provided in the dataset \cite{bik} to extract  the  images  and  prepare  them. Fig. \ref{fig2} shows this process. As we can see, extracted figures from papers contain extra miscellaneous noise that would interfere with the quality of the final training data. Existence of unwanted data in extracted images necessitates the ROI generation process we explain in the following section. 

\vspace{-3mm}
\subsection{ROI Generation}

The main goal of this step of pre-processing is to remove unnecessary information such as labels in the extracted figures to have cleaner images for training process. 
It is needed to get rid of any miscellaneous noise that would interfere with the quality of the final training data, an example of this being text on the images. Also it is important to extract the sections of the images where duplication mostly happens. Based on our observation, duplication is more frequent in images of western blots while it rarely happens in chart, graphs and other type of figures in biomedical articles. Western blots are common in medical or biological publications and are used to study different proteins. It is known that western blots, the parts of the image that duplication occurs mostly,  generally come in the form of rectangular blocks. Therefore, it becomes clear that if anything is not connected to these blocks or found in these blocks it is unnecessary noise. This part of the process is done by removing any part of the image that is not connected to the larger blocks and replaces it with the black mask of that region of the image. 

After this step, the image would ideally have only the sections that have both the unduplicated and the duplicated sections while the rest is a black mask. Fig \ref{fig2} demonstrates the output of this step. As we can see, desired regions are retained while unwanted parts are removed. We use these masks as the inputs of state-of-art manipulation detectors. The results of their detection is discussed in experiment section \ref{secexp}.    

Following the ROI generation step, we segment the western blots knowing that they are rectangular blocks in the output figures from previous step. This is done by finding the contours within the image. This process is made simple because of the fact that only the blocks with sections of interest remained while the rest is a black mask. Once the contours (the blocks of interest) are found, a new image is created using the array of pixels found within each contour. This step segments out the blocks out of the ROI figures which later can be paired and fed to a trained Siamese network to detect whether a given figure of an article has duplication or not.

\begin{figure*}[t]
\includegraphics[width=\linewidth]{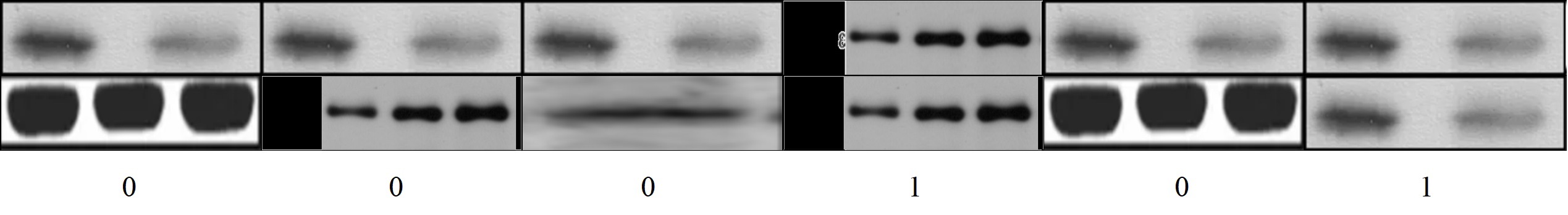}
\caption{Six examples of images pairs fed into our network and their binary output indicating whether or not each pair is copied} \label{fig4}
\end{figure*}


\vspace{-3mm}
\subsection{Duplication Detection using Siamese Network}

A Siamese network is a type of deep learning network that uses two or more identical sub networks that have the same architecture and share the same parameters and weights. The first application of Siamese neural networks was for testing the similarity between two signatures \cite{sig}. Siamese networks are typically used in tasks that involve finding the relationship between two comparable things. Some common applications for Siamese networks include facial recognition, signature verification, or paraphrase identification. Siamese networks perform well in these tasks because of their shared weights. There are fewer parameters to learn during training and they can produce good results with a relatively small amount of training data.



To train the Siamese network, the data must be grouped into pairs of images that are either similar or dissimilar. We prepared both copy-paste pairs and different (unduplicated) pairs from segmented figures. After pairing the images, they are fed into a Siamese network to learn the similarity between images. The ability to learn from very little data made Siamese networks suitable for our task with a few number of images in the dataset. To compare two images, each image is passed through one of two identical sub networks that share weights. For each branch of Siamese network, we use 4 layers of convolution followed by RELU and maxpooling.  The output feature vectors from each subnetwork ($a_{i,j,k}$ and $a_{i,j,k'}$) are combined through subtraction and the result is passed through a fully connected operation to get the final output $a_{i,j,kk'}$, 
where $a_{i,j,kk'}$ is the output of binary classification which determines whether or not $x_{i,j,k}$ and $x_{i,j,k'}$ are  copied.
We use cross-antropy as the loss function for classification task defined as follows:
\begin{equation}
\begin{aligned}
\mathcal{L}_{cls}=\frac{1}{N}\sum_i&\|y_{i,j,kk'}\log(a_{i,j,kk'})\\
&+(1-y_{i,j,kk'})\log(1-a_{i,j,kk'})\|_1,
\end{aligned}
\end{equation}  
where $N$ is number of image pairs. 
Fig \ref{fig2} shows the details of our method for duplication detection. 



\section{Experiments}
\label{secexp}
In this section, we perform the experiments on the dataset we prepared to  investigate  the  efficacy  of  the  proposed  method. We get our results using Segmented Figs while we utilize the Raw Figs and ROI Figs as the inputs for all other state-of-art methods. As shown in Fig. \ref{fig2}, Raw Figs and ROI Figs are output images of figure extraction and ROI generation steps. Segmented Figs are outputs of segmenting and pairing step. We discuss state-of-art methods for manipulation detection in section \ref{secstate}.

\textbf{Dataset.} 
We have access to a dataset manually labeled by Bik \cite{bik} using bounding boxes. To prepare duplicated data for training dataset, we need to use the images with bounding boxes identifying duplicated regions in the figures.  To separate the unduplicated and the duplicated regions, it is important to identify the part of the image that has the multi-colored bounding box.  The image with the bounding box is duplicated and must be refined further meanwhile the ones without the bounding box are not duplicated and no longer need to be changed. To find the duplicated images, we look for any multi-color bounding boxes.  If there is no multi-color bounding box the image is labeled as unduplicated while if the image has them, it is labeled as a copied one.

\begin{figure}[ht]
\includegraphics[width=\linewidth]{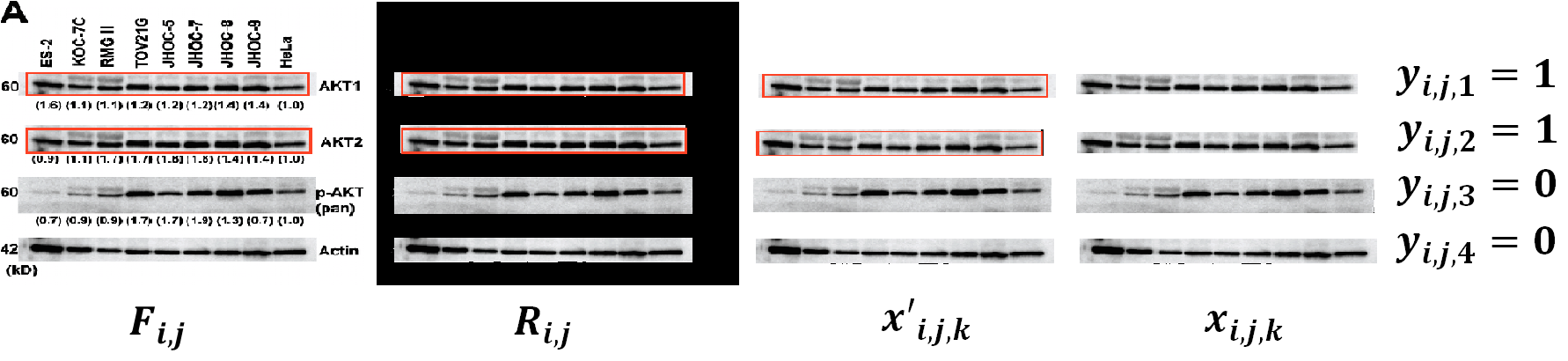}
\caption{All the steps to prepare the dataset including figure extraction, ROI generation, segmentation, bounding box detection and removal} \label{fig3}
\end{figure}

Within the duplicated image directory the next most important step is to remove the bounding box so that only the duplicated section of the image remains. This is once again done by finding the contours of the image and looking for the inner section of the bounding box. And then the pixels within the contours of the image are used to create a new image that has only the duplicated sections. Fig. \ref{fig3} shows the steps for preparing the training dataset. 

\vspace{-6mm}

\subsection{State-of-art Methods}
\label{secstate}
Manipulation detection is a classification problem where the methods focus on identifying whether or not an image or video is manipulated. Copy-paste (duplication) is one type of manipulation which many researchers target for detection. As copy-paste manipulation is abundant in scientific articles , we use two of prominent approaches for manipulation detection; both are pre-trained on CASIA \cite{CASIA} and COVER \cite{cover} datasets containing copy-paste images to classify tampered and clean papers.

\textbf{MesoNet} \cite{mesonet} is a CNN-based network using InceptionNet. The network has two inception modules and two convolution layers with max-pooling, followed by two fully-connected layers.

\textbf{XceptionNet} \cite{xception} uses a deep neural network trained on ImageNet. This architecture is constructed by modifying the inception modules where the depth-wise separable convolution is used. \cite{Rossler_2019_ICCV} transferred it to the manipulation detection task by replacing the final fully connected layer with two outputs. The other layers are initialized with the ImageNet weights. 
\subsection{Quantitative Comparisons}
Table 1  shows  the  classification  accuracy of  different  methods  for detecting duplication in scientific articles. We utilize the output images of figure extraction and ROI generation steps as the input for two state-of-the-art approaches (MesoNet and XceptionNet). For our proposed method, we exploit the output images of segmentation and pairing step. Different steps of dataset preparation are shown in Fig. \ref{fig2}. As we can see from Table 1, XceptionNet and MesoNet have higher accuracy with ROI Figs as the input images in comparison to Raw Figs (shown in Fig. \ref{fig2}). This proves how image pre-processing is effective in the accuracy of duplication detection. To increase the accuracy, we segment out ROI Figs and pair them to feed into the Siamese network. This approach increases the accuracy dramatically. As it is shown in Table 1, our approach achieves 90.86\% of detection accuracy.

The improvement in the accuracy is the result of segmneting the ROI and pairing them as the input for the Siamese network. By doing this, all of the ROI parts are checked to see if there is any similarity between them. Fig. \ref{fig4} shows an example of input and output of the Siamese network. As we can see, the copied western blots are detected as similar ones and the binary outputs are 1 for them while others are 0.  

\begin{table}[ht]
\begin{center}
\caption{Detection accuracy for our proposed method and two other state-of-art methods using Raw Fig. and ROI Fig.}
\vspace{0.2cm}
\begin{tabular}{|l|c|}
\hline
Method & Accuracy\\
\hline\hline
XceptionNet with Raw Fig. & 72.51\\
\hline
XceptionNet with ROI Fig. &76.89\\
\hline
MesoNet with Raw Fig. &62.54\\
\hline
MesoNet with ROI Fig. &68.78\\
\hline
SiameseNet (Ours)&90.86\\
\hline
\end{tabular}
\end{center}
\label{tab:1}
\end{table}
\vspace{-7mm}
\section{Conclusion}
\label{sec:majhead}
In this work we propose a method for detecting the duplication of images within a scientific paper. Our framework emphasizes the importance of image pre-processing steps to prepare appropriate dataset prior to application of deep learning methods. We show that our method leads to $\sim {13\%}$  of improvement in accuracy of duplication detection in comparison to other manipulation detection methods.

\section{Acknowledgement}
We thank Dr. Elisabeth Bik for her critical feedback and advice. We also appreciate her work for preparing corresponding dataset.

\typeout{}
\bibliographystyle{IEEEbib}
\bibliography{refs}

\begin{thebibliography}{10}

\bibitem{bik}
Elisabeth~M. Bik, Arturo Casadevall, and Ferric~C. Fang,
\newblock ``The prevalence of inappropriate image duplication in biomedical
  research publications,''
\newblock {\em mBio}, vol. 7, no. 3, 2016.

\bibitem{Vu}
Linh~Hai Vu, Tsuyoshi Araki, Jianbo Na, Christoph~S. Clemen, Jeffrey~G.
  Williams, and Ludwig Eichinger,
\newblock ``Identification of the protein kinases pyk3 and phg2 as regulators
  of the statc-mediated response to hyperosmolarity,''
\newblock {\em PLOS ONE}, vol. 9, no. 2, pp. 1--13, 02 2014.

\bibitem{Seo}
Jun Seo, Yeonjong Koo, Choonkyun Jung, Song Yeu, Jong Song, J.‐J Kim, Yeonhee
  Choi, Jong~Seob Lee, and Yang Choi,
\newblock ``Identification of a novel jasmonate-responsive element in the atjmt
  promoter and its binding protein for atjmt repression,''
\newblock {\em PloS one}, vol. 8, pp. e55482, 02 2013.

\bibitem{Faneli}
Daniele Fanelli,
\newblock ``How many scientists fabricate and falsify research? a systematic
  review and meta-analysis of survey data,''
\newblock {\em PLOS ONE}, vol. 4, no. 5, pp. 1--11, 05 2009.

\bibitem{ok}
Morten~P Oksvold,
\newblock ``Incidence of data duplications in a randomly selected pool of life
  science publications,''
\newblock {\em Science and engineering ethics}, vol. 22, no. 2, pp. 487—496,
  04 2016.

\bibitem{Acuna}
Daniel~E. Acuna, Paul~S. Brookes, and Konrad~P. Kording,
\newblock ``Bioscience-scale automated detection of figure element reuse,''
\newblock {\em bioRxiv}, 2018.

\bibitem{bucci}
Enrico~M. Bucci,
\newblock ``Automatic detection of image manipulations in the biomedical
  literature,''
\newblock {\em Cell Death and Disease}, vol. 9, pp. 1--9, 03 2018.

\bibitem{sig}
Jane Bromley, James~W. Bentz, Leon Bottou, Isabelle Guyon, Yann Lecun, Cliff
  Moore, Eduard Sackinger, and Roopak Shah,
\newblock ``Signature verification using a “siamese” time delay neural
  network,''
\newblock {\em International Journal of Pattern Recognition and Artificial
  Intelligence}, vol. 07, no. 04, pp. 669--688, 1993.

\bibitem{CASIA}
J.~{Dong}, W.~{Wang}, and T.~{Tan},
\newblock ``Casia image tampering detection evaluation database 2010,'' \url{
  http://forensics.idealtest.org}.

\bibitem{cover}
B.~{Wen}, Y.~{Zhu}, R.~{Subramanian}, T.~{Ng}, X.~{Shen}, and S.~{Winkler},
\newblock ``Coverage — a novel database for copy-move forgery detection,''
\newblock in {\em 2016 IEEE International Conference on Image Processing
  (ICIP)}, 2016, pp. 161--165.

\bibitem{mesonet}
D.~{Afchar}, V.~{Nozick}, J.~{Yamagishi}, and I.~{Echizen},
\newblock ``Mesonet: a compact facial video forgery detection network,''
\newblock in {\em 2018 IEEE International Workshop on Information Forensics and
  Security (WIFS)}, Dec 2018, pp. 1--7.

\bibitem{xception}
F.~{Chollet},
\newblock ``Xception: Deep learning with depthwise separable convolutions,''
\newblock in {\em 2017 IEEE Conference on Computer Vision and Pattern
  Recognition (CVPR)}, July 2017, pp. 1800--1807.

\bibitem{Rossler_2019_ICCV}
Andreas Rossler, Davide Cozzolino, Luisa Verdoliva, Christian Riess, Justus
  Thies, and Matthias Niessner,
\newblock ``Faceforensics++: Learning to detect manipulated facial images,''
\newblock in {\em The IEEE International Conference on Computer Vision (ICCV)},
  October 2019.

\end{thebibliography}

\end{document}